---

# What It Feels Like To Hear Voices: Fond Memories of Julian Jaynes

Stevan Harnad

Chaire de recherche du Canada en sciences cognitives
Institut des sciences cognitives
Université du Québec à Montréal
Montréal, Québec, Canada H3C 3P8

Department of Electronics and Computer Science
University of Southampton
Highfield, Southampton, United Kingdom SO17 1BJ

**Abstract:** Julian Jaynes's profound humanitarian convictions not only prevented him from going to war, but would have prevented him from ever kicking a dog. Yet according to his theory, not only are language-less dogs unconscious, but so too were the speaking/hearing Greeks in the Bicameral Era, when they heard gods' voices telling them what to do rather than thinking for themselves. I argue that to be conscious is to be able to feel, and that all mammals (and probably lower vertebrates and invertebrates too) feel, hence are conscious. Julian Jaynes's brilliant analysis of our concepts of consciousness nevertheless keeps inspiring ever more inquiry and insights into the age-old mind/body problem and its relation to cognition and language.

> "*O, what a world of unseen visions and heard silences,*
> *this insubstantial country of the mind!*"

This is Julian Jaynes's μῆνιν ἄειδε θεὰ Πηληϊάδεω Ἀχιλῆος:

> *"Sing us, O goddess, the wrath of Peleus' son Achilleus,*
> *that countless misfortunes did bring the Achaeans."*

How like Julian to begin his own tome-poem thus, evoking godlessly the genesis of deities and *mens*, in the cadences of the unheard sermons of his father, the Reverend Julian Clifford Jaynes (1854-1922), Unitarian, whom he lost when he was two. Julian will no more have heard those paternal sermons *viva voce* than Homer's epic sung by the sightless poet.

**Aural Tradition.** He would have had to evoke his father's unheard voice in his mind's ear, in reading his *Magic Wells* sermons (Jaynes 1922) (as I did, in reading the copy Julian gave me).

For the voice, I borrowed Julian's own. His diction and even his vocal timbre were also remarkably similar to those of his mother, Clara Bullard Jaynes, whom I heard and spoke to on just one occasion, a phone call from Kepoch, Prince Edward Island, to Julian's remarkably cluttered office on the top floor of Green Hall at Princeton in the mid '70's (that same Green Hall on whose rooftop Jack Shannon – another former student and the one who did the experiments showing the right-hemisphere dominance for perceiving facial expressions -- took that series of photographs of which one haunts us from *Origin*'s dust jacket (Jaynes 1976), and these days all over the web).

Books and papers floor to ceiling, spilling out into a small antechamber adjoining a large classroom, from which, as you approached, you could discern, amidst the spiraling cigar smoke, only the top of a vigorously growing avocado tree that Julian had rescued when it was but a pit, and then the faintly visible silhouette of Julian himself, laboring serenely amidst the papyral debris, his voice occasionally audible, declaiming to his tape-recorder.

**Prisoners of Conscience.** One could also hear Julian's voice, of a Sunday, over the University's radio station WPRB, delivering an occasional guest reading in the service of Princeton's imposing pseudo-gothic chapel, at the invitation of its equally imposing "High-Church" Scottish Presbyterian Dean, Ernest Gordon, whose own memorable sermons were always punctuated at the end with his unforgettably intoned "Wirreld Wutheut And."

Ernest and Julian had more in common than just a pastoral disposition and a fondness for high ritual; both were conscientiously opposed to -- and both had been, each in his own way, prisoners of – war -- World War II in particular, Ernest through the valley of the river Kwai (Gordon 1962), and Julian in a camp far closer to home.

Yet when I arrived in Princeton at the height of the Viet Nam era, Ernest was taking a prominent public anti-war stand, whereas Julian was electing to keep a low profile, eschewing the guru status that would surely have been thrust upon him if he had revealed his past sacrifices; I don't think he was entirely convinced that the war-protesters' conscience ran quite that deep.

**Doctoral Decline.** His mentoring disposition nevertheless saw him become the first master of Wilson College in the new collegiate system at Princeton that he had helped to found. And although he was fond of rite and tradition, he had declined, as we all know, the PhD he earned from Yale (as a protest against the fact that doctorates were conferred for too little, in his view). He said it would not be until he proved worthy of receiving the degree *honoris causa* that he would accept to be endoctored. And of course that day eventually did come, many times over, including the long overdue Yale degree.

But first, after his nondoctorate, he betook himself, with his New England patricians' cadences, to England, and became a Shakespearean actor for a decade, before returning for good to the Academy, eventually to decline tenure at Princeton in favor of a younger family man he felt needed it more.

Apart from that one gesture, Princeton cannot be said to have treated the prophet in its ranks overgenerously, being all too ready to take him at his official (self-denied) rank, which was as a perpetual untenured visiting scholar.

But for the most part I don't think Julian minded, as all he wanted was his office, the library, students, and the occasional seminar he felt inspired to give. And I attended more than a few of those inspired seminars, the harbingers of what is now called "evolutionary psychology" -- and of course the precursors to his theory of consciousness.

**Mirror Neurons.** I also attended many after-seminar palavers at the Annex and the Kings Inn. It's there I met EveLynn McGuinness, his partner at the time, and, as I later learned, a restorative successor to a turbulent earlier relationship of Julian's that had ended tragically. (Julian met EveLynn through joining N.O.W. in the hope of finding a bright and independent-minded counterpart -- and he did indeed inspire EveLynn to abandon her then-secretarial job to become a graduate student, a PhD, a post-doc, and eventually the wife of Professor John Allman at CalTech.)

Julian himself never married, and he was a loner, but I don't think he was lonely, at least not before his last decade. There was one foolish post-EveLynn bout of unrequited love – for someone I shall merely call "Quirk," about a third his age – that inspired some bittersweet poetry that I cruelly mocked, assuming he too perceived the futility as clearly as I did -- something I now regret as having been insensitive and perhaps even injurious on my part. He, it is true, had likewise mocked some romantic melodramas of my own, suggesting -- when I was contemplating creating (under suitably aseptic conditions) a scar on my arm to commemorate "a senseless pain that only I could feel" – that it might perhaps be more dramatic if I were instead to crush a wineglass with my bare hands, "thus." He said it, of course, to dissuade me; but I, of course, went ahead and did the sterile gesture my way, anyway.

There was another time I hurt Julian, this time even more naively, but also more deeply. It was at a dinner at the Annex with visiting faculty, when I spoke far too casually, as if it were self-evident to all, about that affliction that eventually cut Julian's years short, but that was then cutting short only his creative activity and threatening to deprive us all of a sequel to *Origin of Consciousness*. I had fatuously imagined that Julian himself was fully conscious that he was suffering from the self-same malady that had carried away his sister only a few years earlier (for which he was still wearing a black armband on his ancient brown tweed jacket). I suppose I even thought (as he might have thought, with far more justification, with his wine-glass quip) that I might somehow be helping to ward off this tragic fate, with my careless remarks.

**Origin of Language.** Although he owes nothing to me, I owe a great deal to Julian. He became my thesis supervisor for a while, after I had made a shipwreck of my doctoral studies on brain laterality at Princeton. Even he, however, was powerless to ward off my two decades wandering in the doctorless desert – which, unlike him, I was doing through no positive choice of my own. But it was he -- as I was rhapsodizing about the Open Peer Commentary journal, *Current Anthropology* that had inspired me to co-organize the 1975 New York Academy of Sciences conference that re-launched the topic of the Origins of Language after a century long moratorium (a conference to which he too contributed a paper; Jaynes 1976a) -- it was Julian who, when I wistfully whined that it was a pity our field did not have such a peer commentary journal, immediately insisted that I must go and found one. So I did. *Behavioral and Brain Sciences* (of which Julian was Associate Editor, and to which he contributed many memorable commentaries) was the outcome. And that was what kept body and soul together during those long years in the desert.

**Brain Connections.** Julian and I also co-edited a book on *Lateralization in the Nervous System* (Harnad, Doty, Goldstein, Jaynes & Krauthamer 1977), as brain bicamerality began to loom large in his mind. But first there was an interesting Princeton interlude regarding some privileged information that Julian had about the brain of Albert Einstein. Although by now virtually all the pieces of the story are in the public domain, lessons I've learned about discretion and consideration prevent me from putting them all together for you here today. Suffice it to say that Julian saw the brain at a time when almost no one knew where it was, and before the so-called scientific results were published that began trickling in only many decades after Einstein's death in Princeton in 1955. During the time of uncertainty about the brain in the 60's, Julian openly declared that he would be ready to attest to having seen it, and to its whereabouts, in a court of law, if he were ever asked. He was never asked. Meanwhile, Karl Pribram (who is here today and can confirm this!) had since told me that he had generated coronal sections of it on his microtome in the 60's, and that Arnold Scheibel at UCLA was annually displaying one of them to his neuroanatomy students for decades.

Another Princeton/brain connection was Ashley Montagu, the anthropologist and close friend of Julian's. He too participated in the *Origins of Language* conference (Montagu 1976), and brought with him his own personal Neanderthal brain endocasts by way of evidence. Ashley so admired Julian's celebrated obituary of EG Boring that he wanted his own entry in the *International Encyclopedia of the Social Sciences* to be written by the same author. Julian, busy writing *Origins* at the time, proposed that I do it instead. Now Julian is a gifted historian and biographer, whereas I am not (even though I often chided him about being a "historiographer of pygmies," in having decided to devote his talents to chronicling psychologists instead of "real scientists"). But with the experience of writing Ashley's encyclopedia entry behind me, I was able to write an obituary for my other great teacher, DO Hebb, one of the few psychologists one can hardly call a pygmy.

When Julian died, however, I found I could not do it; too close, and too delicate. Perhaps this talk will serve in lieu of that.

Our third brain connection was with Joseph Bogen, and it comes still closer to Julian's bicameral home, because Joe was one of the neurosurgeons who had done the original split-brain series and had gone on to write a number of eloquent and perceptive essays on left-right differences in the brain (Bogen 1995). Joe too contributed to the *Origins of Language* conference and he and Julian immediately hit it off: In fact, Julian and I were each so impressed and exhilarated by the unique persona of Joe Bogen while he was in Princeton that we both found we had caught his drawl as well as his facial expressions for several weeks thereafter. Joe borrowed Ashley's endocasts to make copies in California and Ashley was in a *teddibly* English state of restrained agitation until they were safely returned.

Julian independently created a great admirer in another dear friend, Daniel Dennett, who has been one of the most enthusiastic proponents of the Jaynesian view of consciousness among philosophers (Dennett 1986) (and was the second speaker in this commemorative series in honor of Julian). I will return to this briefly when I get to the real point of my talk. Suffice it to say that the heroic struggles of both Julian and Dan with consciousness originate in their earlier besottedness with behaviorism and that Dan's "Cartesian Theater" had its clear origins in Julian's *Origins*.

I will close these Princeton reminiscences by mentioning Mme Gabrielle Oppenheim-Errera, a grande dame out of the pages of Proust, with a Proustian salon where Julian was a regular invitee, as he was to private tête-à-têtes with Madame O that likewise had a long distinguished pedigree, prior avatars having been the likes of the mathematician Hermann Weyl and the theologian Paul Tillich. The quintessential vignette was when Madame O, then in her high 80's (she lived on, *compos mentis*, till 105) was seated on the carpet[i] chatting with some much younger people, and Julian came up and asked if he could join them: Without missing a beat, she replied cheerfully "Only if you can get up afterwards."

---

**Reflex Machines?** Now to the substance of this talk. In the later years, Julian's dinners at the Annex had scaled down to mostly solitary ones, except when a friend or former student would occasionally join him. As a long-standing vegetarian, I once raised a question about Julian's theory of consciousness that we never did manage, in those dwindling discussions at the Annex, to resolve: How could someone as humane and decent as Julian -- who felt moral and even aesthetic wrongs as acutely as he did, and responded in the empathic and altruistic way he invariably responded -- believe that only humans with language were conscious? Did he, with Descartes, really think that animals were reflex machines that one could kick with impunity, as they were in reality as unconscious as a rock?

Julian gave a reply that is reminiscent of the old joke about the Cardinal who went to the urologist with some awkward symptoms. The urologist immediately recognized that it was syphilis and asked the Cardinal how he had contracted it: The Cardinal replied, indignantly "What a question! It must have been from a toilet seat, of course!" The urologist, knowing that one cannot contract syphilis from a toilet seat, and determined to make the old sinner 'fess up, said: "That is all very well, your Eminence, but I would like you to try to recall as clearly as you can how you might have contracted it, because there are two varieties of syphilis, both fatal if untreated, both curable if treated, but the medicine for the one will not cure the other, and the two medicines are incompatible. Only one is for the variety that is contracted from a toilet seat. So which way would you like to be treated?"

The Cardinal replies, with supreme dignity, "As I am a Cardinal, I got it from a toilet seat, *but I would like you to treat me as if I had gotten it the other way*."

Well, Julian's reply about whether it was alright to kick dogs, since they are unconscious, was rather similar. He replied, that, no, it is not alright to kick dogs -- not because dogs are conscious, which they are not, but because *people* are conscious, and it is hurtful to people to mistreat dogs, both if they witness it and if they do it: If people mistreat dogs, they are more likely to mistreat people in the same way.

It is not that there is no wisdom in that reply. There is moral wisdom. But is Julian right that animals are not conscious?

**The Mind/Body Problem.** We have no choice but to re-enter the question of what "conscious" means, and what it is to be conscious. If nothing else, the condition has a variety of names, both nouns and adjectives: *consciousness, awareness, sentience, subjectivity, intentionality, mind, mental, qualia, experiential state*, *private state, 1st-person state, reflectivity,* etc. etc. So many names, and yet we still have the same old, unsolved problem, usually called the *mind/body problem*, and wrongly attributed to Descartes.

This talk is now about to become a bit heavy-going. Don't worry, though, because it will return to the unresolved question I put to Julian at the very end.

Descartes did suggest that there were two sorts of things, mental and physical, and that these were not the same sort of thing; hence Descartes has also been called a "dualist," which he perhaps was (whatever that means). His Physiology was certainly defective, but its main feature was that it was physical, and that it did not explain the mental, hence the mental was left as some other sort of thing, *sui generis*.

The mind/body problem is: How and why do bodies have minds? or, How and why are some states just physical states whereas other states are also mental states?

But what did Descartes mean by "mental," since that is just one of the many synonyms for conscious? Descartes wrote in Latin; and Latin, unlike French, has a word for mind: *mens* – as in *mens sana in corpore sano*, “a sound mind in a sound body” -- a dualist motto if ever there was one!

**Descartes’ *Cogito*.** But it is not *mens* that Descartes uses when he comes closest to defining mind; it is “Cogito ergo sum.” This is usually translated as “I think therefore I am,” and it was meant to illustrate the one other certainty we have, apart from the certainties of mathematics (which are *necessarily* true, on pain of logical contradiction -- and usually provably so).

Why is “I think therefore I am” as certainly true as “2 x 2 = 4”?

First, it is surprising that the *Cogito* is even a candidate for certainty, because in the world of experience and evidence, apart from mathematics, nothing can be proved to be necessarily true. It is not that things that are not *necessarily* true might not be just simply *true*. Descartes’ *Cogito*, however, is not about truth but about certainty: about what it is that we can be sure about; what is not open to doubt.

And although the laws of physics are almost certainly true, we can’t be sure they are true, or even that they will hold tomorrow. And although the experiential world that we see and hear almost certainly exists, we can’t be absolutely sure, because it could all just be a hallucination or dream. Even when it comes to other people, although it’s almost certain that they too, like us, are conscious, we can’t be sure: they might just be robots who behave as if they were conscious, but are in reality mindless Zombies.

Which brings Descartes to the *Cogito*: If I cannot be certain that other people are conscious, can I even be certain that I myself am conscious?

Descartes’ reply is “Yes”: He puts it in the language of thought: Doubting that one is thinking is itself a case of thinking. So it is true beyond doubt – as necessarily true as that “2 x 2 = 4” -- that I am thinking, if and when I am thinking. While I am thinking I cannot doubt that I am thinking,

Descartes put it in an awkward way. It sounds as if the *Cogito ergo sum* proves more: as if it proved that an “I” exists. But let us defer to Julian Jaynes here, and note that “I” is a rather problematic and theory-laden (even metaphor-ridden) notion. It’s enough if the *Cogito* proves that when thinking is going on, that is as certain and immune to doubt as that P = P.

But that statement alone, put in that formal way, would not have proved that there exist any other certainties than the necessary truths of mathematics. For the statement “If I am thinking, then I am thinking” is no different, formally speaking, from the statement “If I am falling, then I am falling” or even “if the rock is falling, the rock is falling.” It is a logical tautology, necessarily true no matter what, on pain of contradiction.

So that cannot be what the *Cogito* meant either. The *Cogito* was meant to show that there is one *further* kind of certainty, over and above formal tautologies. What is that second kind of certainty, if it is not merely that "I cannot doubt that *if I am thinking then I am thinking*" which is too weak, and it is not "I think therefore *there exists an 'I'*," which is too strong (or perhaps too vague).

**What It Feels Like To Think.** Thinking is a conscious state. When you are thinking, you know you are thinking, because it *feels like* something (Nagel 1974) to think (or to believe, or to know), and whilst you are thinking, you are feeling that feeling (of what it feels like to think that thought).

Forget about so-called "unconscious thoughts" for the moment. They are a legacy from Freud that is most definitely open to Cartesian doubt! And forget also, for now, about Jaynes's valid observation that when we think something, we are not conscious of how our brains arrive at that thought: it is just delivered to us on a platter. That is definitely true; but right now we are not talking about *how* we manage to think, but merely about the fact that we *do* think, and that when we do, we are conscious of it, because it *feels like* something to think.

It's rather hard to express in words *what* it feels like to think. But then it's rather hard to express in words what it feels like to feel *anything at all*: what it feels like to see green, for example, or to hear the sound of an oboe. We can say that green feels soothing or an oboe sounds harsh, but then we have the same problem with what "soothing" and "harsh" feel like. In the end, as with the dictionary definition of a word, once you've exhausted all the definitions of the definitions (Blondin-Massé et al. 2008), and all the similarities of the similarities, you have to rely on the fact that the only way someone else can know what you mean by what X feels like is if they too have felt X.

And X feels like whatever it feels like while we are feeling it – say, seeing or imagining something green, or feeling or imagining a toothache. We can't describe it in words beyond a certain point; we have to rely on the fact that others who have had the same feeling will know what we are talking about.

But the Cogito is not about explaining what one is thinking to someone else. It is about *whether or not one can doubt that one is thinking when one is indeed thinking.* And to show that that is not just a formal tautology, all you need to do is translate it into the more general and revealing language of feeling: One cannot doubt that one is feeling when one is indeed feeling. And that is something each of us knows with certainty, not just as the formal tautology "If one is thinking then one is thinking" but as the feeling itself, whatever it is, and the fact that it is not open to doubt that you are feeling when you are feeling it.

A good example is a toothache: A good Cartesian knows that if I feel a toothache, that does not necessarily mean I have something wrong with my tooth. It might be referred pain from some other region of my body. I might not even have a tooth (in which case it would be "phantom tooth pain"). Or – who knows – I might not even have a body, and the outside world may not exist. None of that is a certainty. But when I do have a toothache – even if I have neither teeth nor a body – I cannot doubt that it *feels like* a toothache. That a feeling feels like whatever it feels like – even if that feeling is an illusion about what exists in the world – is a certainty as indubitable as the necessary truths of mathematics. (Philosophers have also referred to feeling, seeming or appearing as "incorrigible" – and that pertains both to *what* I am feeling and *that* I am feeling.)

(Nor does the fact that you – or even I myself -- might later talk me out of the fact that I felt a toothache a few moments ago change the fact that it felt like whatever it felt like at the time. That sort of retrospective, after-the-fact suggestibility, too, is a bit of spill-over from the legacy of misunderstandings arising from the Freudian hermeneutics of the unconscious "mind". And, before you ask: Hypnotically induced feelings feel like whatever they feel like at the time too, even if the hypnotist has co-opted one's mouth, simultaneously or subsequently.)

**Sentio Ergo Sentitur.** So what the Cogito really means is "Sentio ergo sentitur." Literally, that means "I feel, therefore feeling is being felt." To remove all equivocation about the surplus meaning of the 1$^{st}$ person singular "I," one could even say "sentitur ergo sentitur," accented thus: "*feeling* is being felt, therefore feeling is being *felt*." This is likewise not just a formal tautology (but rather, if anything, has more the flavor of a Kantian "analytic *a posteriori*"), because there is simply no such thing as unfelt or "free-floating" feelings (the very notion makes no experiential sense, any more than one-handed clapping makes logical sense).

Feelings are intrinsically "relational," in the sense that to feel entails both feeling and feeler, passion and patient, object-of-the-feeling and subject-of-the-feeling. Perhaps that's what Descartes meant, or ought to have meant, by the "ergo sum". (And this might be the flip side of Brentano's "intentionality": Brentano took "intentionality" or "aboutness" to be the "mark of the mental" because mental terms like "thinking," "believing," or "knowing" always have a built in *object*, namely, whatever it is that is being thought, believed or known. Well, feelings have a built in *subject*: If something is hurting, someone is also feeling the hurt.)

Now back to Julian and whether dogs are conscious. In his book, he equivocates on the ambiguous, regenerative earthworm -- "The agony of the tail is our agony, not the worm's," he writes, yet I do not for a minute believe that Julian thought that dogs do not really feel pain, but merely behave exactly as if they did (if, say, you cut off their tail). I think he just forgot about that, or set it aside, when he – like so many other thinkers about consciousness -- focused on some of the fanciful metaphors and conceits we have constructed about "consciousness." A good exercise for the reader of *Origin of Consciousness* (or, for that matter, of just about any other treatise, psychological or philosophical, about "mind," "consciousness," and their various other synonyms and paranyms) is to substitute for "conscious" and "consciousness" the words "felt" and "feeling" – and see how much of what is being said still makes sense. For "mind" just substitute "capability of feeling" (and for "mental," once again substitute "felt").

If we substitute “feeling” for “consciousness” then the *content* of consciousness -- whatever it is that we are thinking when we think -- becomes whatever it is that we are *feeling* when we think.

If I say to you “2 x 2 = 4” (and you understand me), then what, exactly, is happening? I am only asking you about what you are conscious of, of course, i.e., what you are feeling, not about any underlying unconscious, hence unfelt, brain processes. I understand “2 x 2 = 4,” and I can think and believe that “2 x 2 = 4,” and I know exactly what that feels like, compared to, say, what it feels like to think, respectively, “1 x 4 = 4,” “1 x 3 = 4” or “kétszer kettő négy.” (The latter is in Hungarian, so you don’t understand it at all, but it also happens to mean “2 x 2 = 4.” Hence, to me, the English and Hungarian versions feel only mildly different, because although they sound different, they mean the same; whereas to you, until I told you what the Hungarian version meant, they would feel very different, because one sounded both acoustic *and* meaningful and the other just sounded like nonsense syllables).

Now I don’t want to disguise from you the fact that I have done something very unorthodox here: If I had simply played you the sound of an oboe, and asked you what it feels like to hear that sound, you may have some trouble putting it into words, but you would have no trouble understanding what I meant by “what it feels like to hear that sound” (apart from your slight preference for saying, instead “what it *sounds* like to hear that sound”) – but please let me use my generic term “feeling” for sensing in any of the sensory modalities, because I will need it for the so-called “amodal” case that makes what I am saying now so controversial.

**What It Feels Like To Mean.** I am talking about *the feeling of meaning*: What it feels like to understand, versus not understand; or what it feels like to understand this rather than that. Same thing for what it feels like to think this or that, or what it feels like to mean this or that (as in, “No, my intended meaning was not this, but that.”)

In the case of the oboe, we have no trouble with saying that the content of what we are conscious of when we hear an oboe, and then hear a flute, is different, in that it feels like this to hear an oboe and it feels like that to hear a flute.

Now when it comes to spoken words and their meanings, there is, first, their sound, which differs from utterance to utterance. For example, it sounds different to say “2 x 2 = 4” versus “1 x 4 = 4,” “1 x 3 = 4” or “kétszer kettö négy”. But, in addition to the difference in sound, there is a difference in meaning between “2 x 2 = 4” and “1 x 4 = 4,” though they both are (and feel) true, and an even bigger difference from “1 x 3 = 4” (because it is and feels false). As a consequence, what it feels like to think and understand these three strings of symbols differs not just in what each sounds like, but also in what each means. (Moreover, for a Hungarian/English bilingual, thinking “2 x 2 = 4” and “kétszer kettö négy” differs only in what they sound like, but not in what they mean.)

Why is this so controversial? Because the usual view is that *a difference in meaning is not like a sensory difference, hence it is not a felt difference*. Words mean what they mean irrespective of *what* we feel, or even *whether* we feel. Their meaning is wide -- based on social agreement amongst language-users about what we will name what, plus whatever turns out to be true about the things we name. "Cat" means cat and not dog irrespective of how I feel about it, unless I insist on taking a Humpty-Dumpty view of language (insisting that words mean whatever I *choose* them to mean -- and in that case Wittgenstein has a thing or two to say about the impossibility of such a private language, because there is no longer any criterion, public or private, for what is *right* and what is *wrong*).

And I reply, yes, yes, that is all true, but we are not talking here about *wide meaning* in the case of the *Cogito*. Wide meaning concerns what language communities agree to call things and what turns out to be true about those things in the big wide world. But we (with Descartes) are talking about *narrow meaning*, in the head.

When I say something to someone else, I had better get my wide meaning straight, otherwise I may be saying something false or even unintelligible in relation to what others mean and what is true in the world. But if we are talking about narrow meaning in my head, the only one I am answerable to is myself, just as I (or, for that matter, a nonlinguistic ape) am only answerable to myself about what toothache feels like. When I have a toothache, it feels like something unpleasant happening *here*, even if there is in reality no tooth here, and even if there isn't even a "here" here! (Julian is absolutely right about the physical location of feeling: Feelings may feel-like they are somewhere in space, but that doesn't mean they really are where they feel-like they are.)

**Feeling-Space.** Feeling-space is qualitative, like the color-spectrum, and it is ruled by something similar to the JNDs of psychophysics, the "just noticeable differences": Two sensations are different only if we can tell them apart: if they are too small to tell apart – if their difference is below the psychophysical threshold for a JND -- then they are the same sensation. But the reason I said feeling differences are merely *similar* to psychophysical JNDs rather than identical, is that JNDs of course have an objective reference point: the external stimulation, the input. The psychophysicist can take a physical input signal – say, different levels along the one-dimensional continuum of intensity in the sound of an oboe -- and can reduce the difference in intensity between two levels until you can no longer hear any difference. That is the JND. Above the JND threshold, the oboe sound intensity continuum varies in how loud it feels, and if you add more dimensions of variation, you get pitches and timbres, and all the possible sounds of an orchestra (or, if we consider vision instead of hearing: the color palate of a painter) – all further dimensions in feeling-space.

Well, I am suggesting that meaning-space – dare I call it "semantic space"? – is rather like that too, and likewise a part of overall feeling-space: Meaning differences, too, constitute a felt, multidimensional qualitative space that is not only similar to sensory psychophysical space, but coupled with it – at least inasmuch as we think in words (and hence also the sounds of the words) and we think about the things in the world that words refer to, hence what those things look, sound and taste like.

**Consciousness and Language.** Now if this theory is correct—if not just sensory differences but also semantic differences are indeed felt differences -- then we must take a view of both language and consciousness that is different from the one taken by not only Julian Jaynes, but also most philosophers of language and mind, even those that disagree with Julian's theory of the nature and origin of consciousness and its relation to language.

The standard view is that sensation is one thing, language is another. Whoever is inquiring about the origins of feeling is inquiring about the origins of sensations, whereas whoever is inquiring about the origins of language is inquiring about something else, something grounded in and drawing upon feeling, to be sure, yet something that is also somehow autonomous. And it is precisely in this "somehow autonomous" that most of the disagreement and equivocation resides.[ii]

Julian misremembers Locke's *Nihil est in intellectu quod non antea fuerit in sensu* in paraphrasing it as "There is nothing in consciousness that is not an analog of something that was in behavior first." Locke is referring explicitly to *sensation* here, not to *behavior*; and Julian is here betraying the latent behaviorist in him, against which he battled (and over which he triumphed) throughout his career.

**Explaining Behavioral Capacity.** Let us cut to the quick: Psychology and "cognitive science" would have been a lot simpler if there had been no consciousness at all: If organisms had indeed merely been Darwinian survival machines, with brains that are able to behave adaptively, so as to achieve reproductive success. Then the only scientific task would be to "reverse-engineer" organisms' powerful behavioral *capacities* (as Dan Dennett -- who likewise still harbors cryptobehaviorist tendencies -- would rightly put it) in order to explain how the brain manages to generate those capacities.

(Of course the real behaviorists never really explained -- or even tried to explain -- anything at all. They thought schedules of rewards and punishments would account for whatever we do – not realizing that the real problem was to explain what it was in organisms that made it *possible* for schedules of rewards and punishments to shape organisms into doing what they do.)

But if we loosely call "cognitivism" that successor to behaviorism that no longer rejects, as the behaviorists did (in an over-reaction against introspectionism) the attempt to explain what is going on unobservably inside the head, in order to explain the causal basis of our behavioral capacities, then *all we can ever expect from cognitivism is a causal explanation of how we are able to do all the things that we are able to do*.

**The Turing Test.** And explaining all the things we can do also includes explaining all the things we can *say*, i.e., all of our linguistic capacity. It was for this reason that Turing's famous challenge to (what would eventually be called) cognitive science was to engineer a system that had our full linguistic capacity (Turing 1950): To pass the Turing Test, the candidate system must be able to communicate with people by

email in such a way as to be indistinguishable from a real (lifelong) pen-pal (Harnad 2007).

It is unlikely that any human-designed candidate could pass the Turing Test if it had only linguistic capacities. It also would have to have the sensorimotor (i.e., robotic) capacities to do everything else that we can do, even if those robotic capacities are not directly tested by the original (linguistic) version of the TT. This is true for much the same reason that (as we noted earlier) word meanings cannot all be learned from dictionary definitions alone: Some of them first have to be grounded in sensorimotor (i.e., robotic) experience (Cangelosi & Harnad 2001).

This is (or ought to be) why Locke said *Nihil est in intellectu quod non antea fuerit in sensu*. “Nothing is cognized that was not first sensed.” Language is not an island unto itself.

But lest you conclude that adding sensorimotor robotic grounding to linguistic capacity and providing a full causal explanation of both these behavioral capacities (perhaps even one that was explicit enough to successfully pass the full-blown TT) would solve the mind/body problem and explain consciousness, I’m afraid it is not so.

**The Feeling/Function Problem.** For a successful TT robot that could pass indistinguishably as one of us (for a life-time, if need be) might still be an unconscious Zombie (Harnad 2003) – of the kind that Julian imagined we *all* were in the Bicameral era – except that the TT robot (being indistinguishable from us) would be post-Bicameral, just like us, and yet it could still be a Zombie. And that is because the missing ingredient (if it was indeed missing) would not be – as it was for Julian – a certain way of narratizing about the mind. (Our TT robot, even if it were a Zombie, would have to be able to narratize as we do too, both externally and internally.) It would be the fact that the TT-robot failed to feel (if it indeed failed to feel -- which is yet another thing of which we can have no Cartesian certainty, one way or the other, for anyone other than ourself) that made it an unconscious Zombie, hence no explanatory solution to the mind/body problem.

Now we come back to the unresolved question of why Julian, this most humane of men, would never kick a dog: We must now ask the same question about the hypothetical Turing-Test-passing robot: Would Julian kick that? If I had revealed to him that I myself consisted of transistors and effectors, crafted in the robotics lab at McGill only a few years before I met him, rather than engendered in Hungary at the close of World War II, would he then have felt that the only reason he would not kick me would be that it would set a bad example for post-Iliad humankind?[iii]

No, I think Julian’s especially sensitive “mirror neurons” (Gallese & Goldman 1998) – those lately discovered brain-cells that are active only when I and someone else are in the same mental state – would restrain him from kicking me as surely as Ernest Gordon’s mirror neurons restrained him from kicking his tormenters in the valley of the Kwai.

Remember that Descartes was inquiring not about truth but about certainty. We cannot be certain that others feel – only almost certain. But by the very same token, we cannot be certain that others do *not* feel, whether or not they can speak, to tell us. That – along with our mirror neurons – is why we give the benefit of the doubt to the dog (and our fellowmen). Julian wrote about the problem of "animate motion" (Jaynes 1973). Well, animal agony is as apparent to our "mirror neurons" as animate motion is: We feel what others feel, and Julian felt it acutely. Not only would he never have kicked a dog, or a Turing-scale robot that was otherwise like the rest of us, but he would not have kicked a bicameral Greek either – for it *feels like* something, whether to think or to hear voices.

We are not just narratizing Darwinian survival-machines: Turing Zombies. Each of us is a feeling creature. And it is this private fact -- to which one each of us is privy, putting it beyond the reach of all Cartesian doubt -- that has driven some of us to invent omnipotent deities and immaterial souls; and led others simply to admit that we have here an unsolved, and perhaps insoluble explanatory problem.

The mind/body problem – or, as I prefer to call it, the feeling/function problem – was at the heart of Julian's heroic attempt to explain consciousness. It had also been at the heart of the behaviorists' craven attempt to deny it.

Julian Jaynes has not solved the problem of consciousness; but his world of unseen visions and heard silences has inspired countless others to explore ever more deeply, this insubstantial country of the mind.

---

[i] *Chez quelqu'un d'autre*, of course. *Chez les Oppenheim-Errera* one did not carry on *parterre*, except perhaps in the case of Lord and Lady Russell on one notable occasion.

[ii] **"2nd Order Consciousness."** A fundamental error made by many philosophers and non-philosophers alike is to equate consciousness with so-called "2nd order consciousness" and to argue that there is some sort of profound difference between "merely" being aware, and "being aware of being aware" -- with the latter being the "real" problem of consciousness.

As soon as one replaces all the many synonyms for "consciousness" by the far less equivocal term "feeling," it becomes apparent that "2nd order consciousness" -- which simply becomes "feeling that one is feeling" -- is itself just a feeling, like any other, with no especially privileged status.

We would not single out feeling tired or feeling that it is sunny today as having any special status among the many possible feelings we can have (or, to put it another way: among the many possible contents of feelings, the many things that we might feel). Then why especially single out feeling that we feel? There are many banal counterparts: Looking at Tom looking at Dick. That feels like something too. And it doesn't stop there, for we can feel what it feels like to look at Tom looking at Dick looking at Harry. (Are these to be feelings of a still higher order?) And we can also feel what it feels like to look at our image in a mirror that is facing a mirror behind us, hence what it feels like to look at ourselves looking at ourselves looking at ourselves, and so on, as far as you can see and count.

There are similar effects that are singled out as special in speech and discourse theory, in the form of the presuppositions of speech acts: "I know that you think that I believe that you want..." Yes, it feels like something to perceive or assume or believe something like that; but only in the same sense that it feels like something to see green or to understand that "2 x 2 = 4." Or to feel what it feels like to understand a sentence like "This is the cat that chased the rat that ate the malt that lay in the house that Jack built."

The problem of "merely being aware" -- i.e., the problem of feeling -- is not a "lesser" problem of consciousness. It *is* the problem of conscious, the mind/body problem (and the *only* mind-body problem). Explain how and why we can be "merely aware" (i.e., "merely feel") and you've solved the whole problem. Enter instead the hermeneutic hall of mirrors generated by "higher order consciousness" and you are only in an amusement park fooling yourself -- as those who like to intone the mantra of "reflexive" versus "reflective" awareness do: For of course "reflexes" are no kind of awareness at all, if the system does not feel; they are merely movement. And if they are indeed felt actions -- whether automatic or deliberate -- then they are already infected with the full-blown-problem of consciousness.

As to whether we really cause our actions (i.e., really do what we want because we feel like it -- or it merely *feels like* it), well, that's just the flip side of the mind/body problem (Libet 1985).

[iii] Or would my “narratizing” capacity somehow constitute “proof” that I felt? How? Why?